\documentclass[journal]{IEEEtran}
\ifCLASSINFOpdf
\else
\fi

\usepackage{amsmath,amssymb,amsfonts}
\usepackage[usenames,devipsnames,table,xcdraw]{xcolor}
\usepackage{etoolbox,pgf}
\usepackage{xurl}
\usepackage{csquotes}
\usepackage{flushend}
\usepackage{amsmath}
\usepackage[ruled]{algorithm2e}
\usepackage{tabularx}
\usepackage[T1]{fontenc}
\usepackage[super]{nth}
\usepackage{multirow}
\usepackage{hyperref}
\usepackage{graphicx}
\usepackage{booktabs}
\usepackage{pifont}
\usepackage{textcomp}
\usepackage{comment}
\usepackage{url}

\usepackage{xskak}

\definecolor{1}{HTML}{efa97e}
\definecolor{2}{HTML}{e7736f}
\definecolor{3}{HTML}{7abc81}
\definecolor{4}{HTML}{f0e891}
\definecolor{5}{HTML}{eee791}
\definecolor{6}{HTML}{efa97e}
\definecolor{7}{HTML}{b5d389}
\definecolor{8}{HTML}{f6ea92}
\definecolor{9}{HTML}{f3c686}
\definecolor{10}{HTML}{fae58f}
\definecolor{11}{HTML}{ec9378}
\definecolor{12}{HTML}{f9eb92}
\definecolor{13}{HTML}{dde18e}
\definecolor{14}{HTML}{e7736f}
\definecolor{15}{HTML}{e5e48f}
\definecolor{16}{HTML}{fae58f}

\begin{document}
%
\title{Neural Network-based Information Set Weighting for Playing Reconnaissance Blind Chess}
%
%
%

\author{Timo~Bertram,
        Johannes~F\"urnkranz,
        and Martin M\"uller
\thanks{Timo Bertram and Johannes F\"urnkranz are with the Institute for Application-Oriented Knowledge-Processing at the Johannes Kepler University Linz, Austria and affiliated with the LIT Artificial Intelligence Lab (e-mail: tbertram@faw.jku.at, juffi@faw.jku.at)}
\thanks{Martin M\"uller is with the Dept. of Computing Science and the Alberta Machine Intelligence Institute at the University of Alberta, Canada (e-mail: mmueller@ualberta.ca)}
\thanks{Manuscript received 12.2023; revised 07.2024.}}

%
%

\markboth{IEEE Transactions on Games,~Vol.~?, No.~?, Month~Year}%
{Bertram \MakeLowercase{\textit{et al.}}: Learning to Play Reconnaissance Blind Chess}
%


\maketitle


\begin{abstract}
    In imperfect information games, the game state is generally not fully observable to players. Therefore, good gameplay requires policies that deal with the different information that is hidden from each player. To combat this, effective algorithms often reason about information sets; the sets of all possible game states that are consistent with a player's observations. While there is no way to distinguish between the states within an information set, this property does not imply that all states are equally likely to occur in play. We extend previous research on assigning weights to the states in an information set in order to facilitate better gameplay in the imperfect information game of Reconnaissance Blind Chess. For this, we train two different neural networks which estimate the likelihood of each state in an information set from historical game data. Experimentally, we find that a Siamese neural network is able to achieve higher accuracy and is more efficient than a classical convolutional neural network for the given domain. Finally, we evaluate an RBC-playing agent that is based on the generated weightings and compare different parameter settings that influence how strongly it should rely on them. The resulting best player is ranked 5\textsuperscript{th} on the public leaderboard.
\end{abstract}

\begin{IEEEkeywords}
Siamese Neural Networks,
Information Sets,
imperfect information Games, 
Reconnaissance Blind Chess
\end{IEEEkeywords}

%
\IEEEpeerreviewmaketitle

\newcommand{\set}[1]{\ensuremath\mathcal{#1}}
\newcommand{\vect}[1]{\ensuremath\mathbf{#1}}

\newcommand{\I}{\set{I} }
\newcommand{\Icard}{|\I|}
\newcommand{\F}{F} 
\newcommand{\e}{e} 
\newcommand{\Fstar}{F^*}
\newcommand{\anchor}{a}
\newcommand{\positive}{p}
\newcommand{\negative}{n}
\newcommand{\Obs}{\set{O}}
\newcommand{\ra}[1]{\renewcommand{\arraystretch}{#1}}
\newcommand{\triplet}[3]{\ensuremath \langle #1,#2,#3 \rangle}

\section{Introduction}
%
%
%
%
\IEEEPARstart{G}{ame AI} has achieved superhuman performance in many classic, fully observable games such as chess \cite{campbell2002deep}, Go \cite{silver2016mastering}, and backgammon \cite{tesauro94TDgammon}, as well as in imperfect information games like Poker \cite{brown2019superhuman}. Imperfect information games, in contrast to fully observable ones, feature a game state that is only partly visible to a player, which increases their difficulty and makes them an attractive area for artificial intelligence research. For example, playing imperfect information games requires probabilistic mixed policies to avoid being predictable and therefore exploitable by other players. As a formal model of player's uncertainty, game theory for imperfect information games \cite{von1947theory} clusters all states which one player cannot distinguish into an \textit{information set}. However, while players can generally not identify the true game state, states within an information set are not equally likely in practice -- sophisticated play by agents should reach some states more frequently than others. Furthermore, opponent modelling from observing past behaviour can give strong clues about the likelihood of their future actions. 

\break
In this work, we employ Siamese neural networks to learn a function that maps an information set $\I$ of game positions to a weight distribution, which translates to the probability of each state being the true game state. As a proof-of-concept of the utility of such a distribution, we create a simple player for the imperfect information game \textit{Reconnaissance Blind Chess}, which plays the game based on the best perfect information responses on the boards of the information set, weighted by the obtained distribution.

\vspace{5pt}
First, we briefly introduce Reconnaissance Blind Chess in Section~\ref{sec:RBC}. We then formalise and motivate the approach of reducing the evaluation of an imperfect information state to a weighted evaluation of perfect information states in Section~\ref{sec:problem}. As a proposed method for this, we introduce our approach with Siamese neural networks and relate it to previous work in Section~\ref{sec:overview}. The experimental setup including data preparation and the training process is described in Section~\ref{sec:experiments}. In Section~\ref{sec:cnn}, we evaluate our Siamese network approach by comparing it to a conventional baseline; a convolutional neural network that directly estimates the probability of a board state given an information set. Lastly, we test the quality of our idea in a predictive setting (Section~\ref{sec:Evaluation}) and as an integral part of an actual agent (Sections~\ref{sec:rbc_agent} and~\ref{sec:playing_performance}).

While the key ideas of information set weighting via Siamese neural networks have already been introduced in our prior work \cite{Bertram_Weighting_Information_Sets}, this extended version of the \textit{IEEE Conference on Games 2023} conference paper elaborates on it in several aspects:

\begin{itemize}
    \item We add a more detailed motivation of the problem and discussion of the approach.
    \item Section~\ref{sec:related_work_imperfect_info} offers a more elaborate review on related algorithms for imperfect information games.
    \item Section~\ref{sec:cnn} introduces a new baseline to the paper. By comparing to a conventional neural network, we show that our contrastive model is more computationally efficient in testing and achieves higher overall performance.
    \item Section~\ref{sec:rbc_agent} adds a more detailed account on how to use the weighting scheme in gameplay.
    \item Section~\ref{sec:playing_performance} investigates a temperature parameter responsible for the smoothness of the weight distribution, i.e. setting the optimism of the agent. 
\end{itemize}

\clearpage

\begin{figure*}[t]
    \centering
    \includegraphics[width = \textwidth]{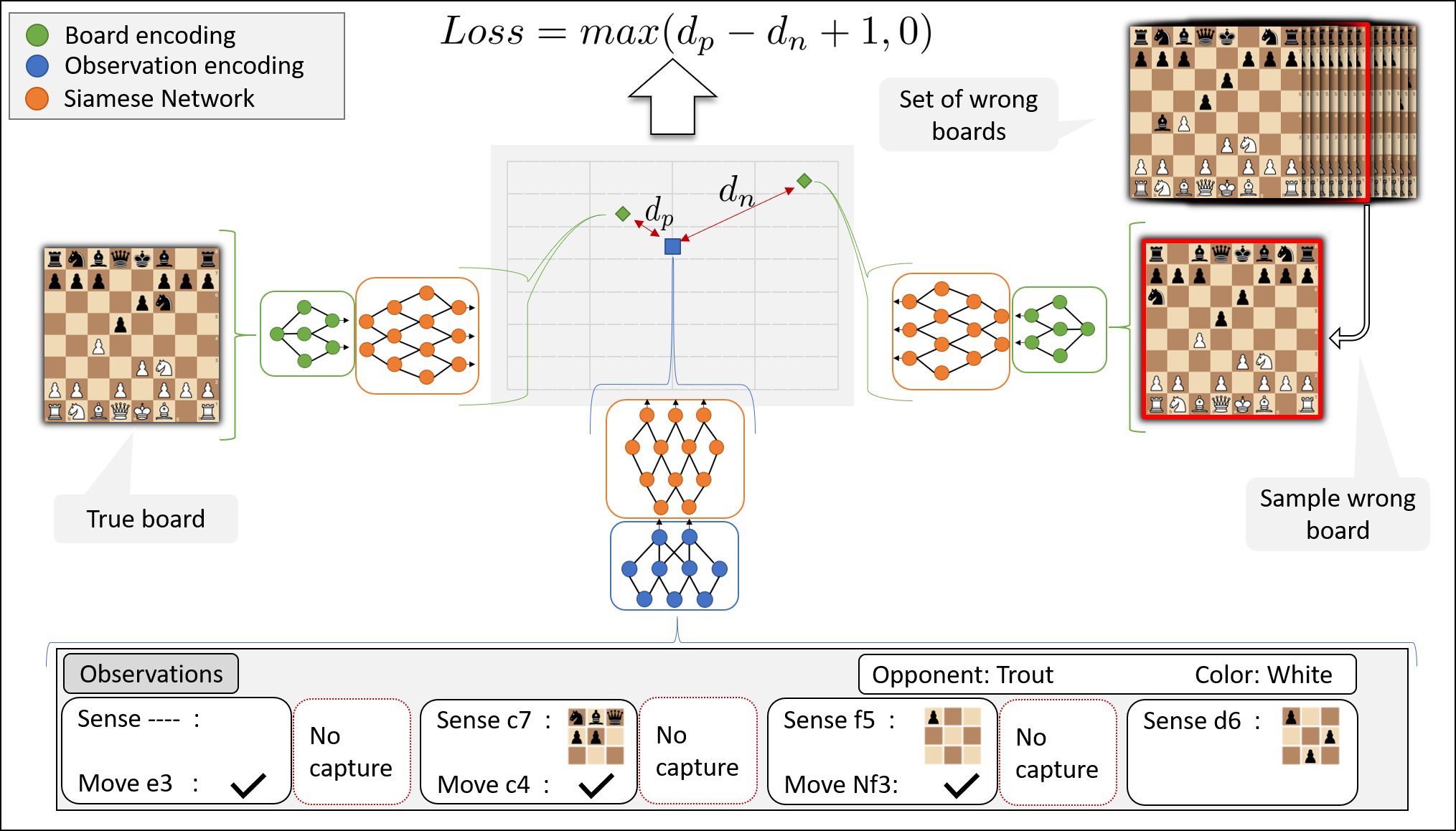}
    \caption{Schematic overview of using a Siamese neural network for weighting \textit{RBC} board states in an information set. The observation is preprocessed by an observation encoding network, while both the real board and a sampled incorrect board are preprocessed by a board encoding network. Next, all are input into the Siamese network. The distances of the outputs in the embedding space model the probabilities of boards being true given the observations.}
    \label{fig:scheme}
\end{figure*}

\section{Reconnaissance Blind Chess}
\label{sec:RBC}

\textit{Reconnaissance Blind Chess} (RBC) \cite{gardner2020first} is an imperfect information variant of classical chess. While the game pieces start in the same arrangement as in classical chess, players do not receive full information about their opponent's moves and are generally unaware of the exact configuration of their opponent's pieces. A turn of Reconnaissance Blind Chess consists of four parts:
\begin{enumerate}
    \item First, the player receives a limited amount of information about the latest opponent move: If the opponent captured one of the player's pieces, the player is only notified about the square where the capture occurred. For non-capture moves, no information is obtained.
    \item The player is allowed to \textit{sense} a 3×3 area of the board, which reveals all pieces in that region. Players never receive any information about the opponent's senses.
    \item With some minor differences, a player decides on a move as in conventional chess. Due to the missing information, players frequently cannot decide whether a move is legal.
    \item Finally, the player receives information about whether their chosen move succeeded. 
\end{enumerate}

One turn of \textit{RBC} includes two separate actions --  \textit{sensing} and \textit{moving}. The result of the \textit{sense} is obtained immediately, and the \textit{move} is chosen with that information. Due to the imperfect information nature of \textit{RBC}, some rules differ from normal chess:
\begin{itemize}
    \item Players can attempt illegal moves. \enquote{If a player tries to move a sliding piece through an opponent's piece, the opponent's piece is captured and the moved piece is stopped where the capture occurred} \cite{RBC}. Other illegal moves are converted into a pass.
    \item Players are not notified when their king is put in check. They are also allowed to move their king into check, and castle through and out of checks. There are no draws by stalemate, where a player is unable to move without putting their king into check.
    \item Most games end by capturing the opponent's king. Rare draws result from 50 consecutive turns without game progress, as in classical chess.
\end{itemize}

These rules imply the following properties of \textit{RBC}:

\begin{itemize}
    \item A player receives enough information to always perfectly know the placement of their own pieces.
    \item Because of perfect recall, it is always possible to compute the current information set. However, the size of this set can become large, especially when sensing is ineffective or when players play erratically.
    \item Speculative moves and aggressive strategies that directly launch a concealed attack towards the opposing king, which would be self-destructive in classical chess, become a major factor in \textit{RBC}. Such strategies can win quickly if the opponent does not recognise and defend against them.
\end{itemize}

With these changes in mind, we now introduce the problem setting formally.

\section{Problem Statement}
\label{sec:problem}

\subsection{Imperfect Information Games}
In an extensive form imperfect information game, a game state $\e$ captures all information of the current situation, including the private information of all players. An information set $\I$ for a player is a set of states $\{\e_1, \e_2, \dots,\e_{\Icard}\}$ which are indistinguishable from that player's perspective. In the case of RBC, this is the set of boards, i.e. piece configurations of the opponent, that are consistent with the information a player has received throughout the game. Formally, the private observation history at player's turn $t$ with $\Obs_t = (o_0,\dots,o_t)$ implicitly defines their information set $\I_t$. In Figure~\ref{fig:scheme}, the observation history of the current player is depicted at the bottom, consisting of the current position of one's pieces, the name of the opponent, and the history of previous senses and moves, including information about whether the moves were successful, and whether pieces were captured by opponent's moves (see Table~\ref{tab:observations}). Of the boards in the information set, one is the correct board $\positive_t \in \I_t$, whereas all other boards $\negative_{t,i} \in \I_t \setminus \{\positive_t\}$ differ from the true hidden game state. In Figure~\ref{fig:scheme}, the correct board $\positive_t$ is shown on the left, whereas the boards $\negative_{t,i}$ are shown in the upper right corner.

\subsection{State-of-the-Art in Imperfect Information Games}
\label{sec:related_work_imperfect_info}

As there are numerous algorithms used in imperfect information games, we can not give a complete overview of all of them. However, we briefly review a subset of the current state-of-the-art approaches for the most popular games.

In \textit{Poker}, various forms of counterfactual regret minimization \cite{zinkevich2007regret}, such as $CFR^+$ \cite{CEPHEUS}, Monte Carlo counterfactual regret minimization \cite{burch2012efficient, brown2018superhuman,brown2019superhuman}, or \textit{DeepStack} \cite{Moravcik2017} reign superior. Algorithms such as ReBel \cite{brown2020combining} use the same general framework but add reinforcement learning to CFR. Fictitious play-based algorithms \cite{heinrich2016deep} use deep reinforcement learning from self-play, but have not yet reached the same performance level as the previously mentioned algorithms. However, newer self-play reinforcement learning approaches \cite{zhao2022alphaholdem} are more light-weight than CFR and have shown promising results.

\textit{Stratego} has similar imperfect information components as Poker but requires long-term strategy. Here, the current state-of-the-art algorithm, \textit{DeepNash} \cite{perolat2022mastering}, uses deep reinforcement learning and self-play. It does not use search due to the high difficulty of estimating the opponent's hidden information. 

Commercial card games such as \textit{Hearthstone} and \textit{Magic: The Gathering} inherently feature a vast amount of imperfect information. So far, few complete agents for those games exist, and the majority of research has been focused on pre-gameplay tasks such as card balancing and deck building. Recently, one of the first human-level agents for \textit{Hearthstone} \cite{xiao2023mastering} was constructed based on fictitious self-play and Monte Carlo Tree Search. Generally, MCTS can be applied to imperfect information games with some adaptions \cite{cowling2012information}.

Faster, commercial, non-sequential games feature imperfect information through obstruction of the visible game space. In games such as \textit{Starcraft} and \textit{Dota} \cite{berner2019dota, vinyals2019grandmaster}, large-scale deep reinforcement learning is typically used, and hidden information is regarded as an intrinsic part of the environment. Due to the fast reaction needed, such approaches typically do not use search and rather directly sample from the policy output of the networks to generate actions.

Finally, we want to emphasise that our method is not intended to be used as a full-scale imperfect information player. While we show in Section~\ref{sec:playing_performance} that it is straightforward to do that, we focus on the task of learning a distribution over the states in the information set rather than their explicit use in a game. More work is required to evaluate how exactly an accurate distribution over the information state should be used in practice.

\subsection{Weighting Information Sets}

In several imperfect information games, remarkable performance has been achieved by basing the imperfect information gameplay, whether implicitly or explicitly, on perfect information evaluations of states in an information set. Bl\"uml et al. \cite{AlphaZeStar2} show that proficient gameplay can be achieved in imperfect information games with an approach akin to AlphaZero \cite{silver2018general}, in which they fuse policies across different states. Bertram et al. show that one can directly formulate a policy using the information set as an input \cite{bertram2022supervised}, and many Monte Carlo Tree Search-based methods individually evaluate states of an information set \cite{browne2012survey} before combining the results. Additionally, strong programs in \textit{RBC} usually rely heavily on classical engines for evaluating chess positions \cite{gardner2020first,perrotta2022second,gardner2022machine}, using the idea of approximating the public state evaluation with the expected value of the states in the information set. 

The idea behind such approaches is to approximate the evaluation $g(m,s)$ of a move $m$ in an imperfect information state $s$ with the weighted sum of the evaluations of the resulting perfect information game states in the information set $\I_s$:

\begin{equation}
g(m,s) \approx \sum_{e\in\I_s} w_{e,s} \cdot f(m,e) 
\label{eq:eval}
\end{equation}
where $f(.,.)$ is a perfect information evaluation of a move in a state of the information set and 
$w_{e,s}$ is the weight assigned to that position. In \textit{RBC}, $f(.,.)$ is often the evaluation of a chess position using a conventional chess program such as Stockfish. A simple weighting can be achieved by considering all states $e$ in the information set as equal ($w_{e,s} = \frac{1}{|\I_s|}$), as has, e.g., been done in \cite{AlphaZeStar2} with surprisingly strong performance. Other choices are to only use the most likely state and ignore all others ($w_{e,s} = [\![ e = \arg \max w_{e,s} ]\!]$) \cite{Bertram_Weighting_Information_Sets} or to interpret the weights as a probability distribution ($w_{e,s} = P(e|s)$), in which case Equation~\eqref{eq:eval} computes the expected value of the given move over all states in the information set.

\begin{figure}[b]
    \centering
    \includegraphics[width = 0.6\columnwidth]{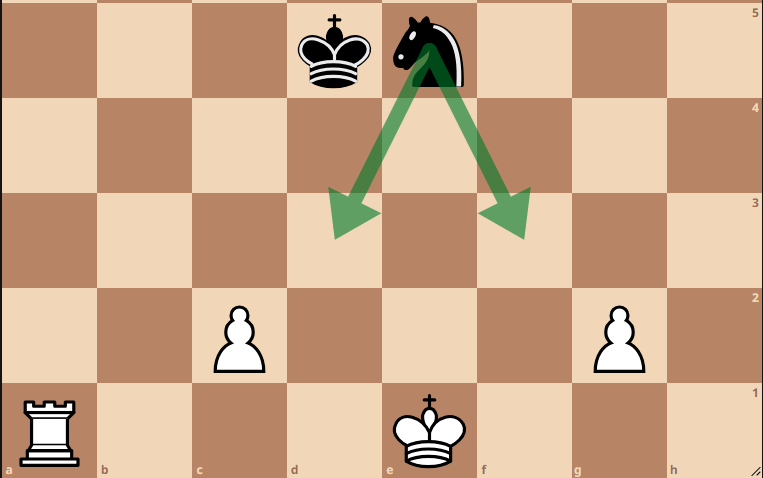}
    \caption{Example black-to-move position, where the best overall move is suboptimal on either board.}
    \label{fig:position}
\end{figure}

It is important to acknowledge the limitations of basing imperfect information policy fully on perfect information evaluations. An example of such a case is shown in Figure~\ref{fig:position}: Assume that white knows that the Knight was previously on \textbf{e5} and now moved to a different square, threatening to take the King from either \textbf{d3} or \textbf{f3}.\footnote{For the sake of conciseness, we ignore irrelevant moves.} Thus, the information set consists of two positions, $e_{1}$ in which black has moved \variation{1...Nd3}, and $e_{2}$ in which black has moved \variation{1...Nf3}. In either position, white can take the Knight with moves $m_1 = $\variation{2.cxd3} and $m_2 = $\variation{2.gxf3} respectively, which in each case is the best move on that particular board. However, because we have to decide on a move which is evaluated in both states, playing either move will result in winning in one option and losing in the other one where black can take the King. Thus, the evaluation of either move $m_1$ or $m_2$ can be computed as $g(m_i,s) = w_{m_i,e_1} \cdot 1 + w_{m_i,e_2} \cdot -1$, which is close to $0$ with the assumption that $w_{m_i,e_1} \approx w_{m_i,e_2} \approx 0.5$. However, the best overall move for the imperfect information state, which is completely separate from the best move for either perfect information state, would be to simply move the king to a safe square, e.g., $m_3 = $ \variation{2.Kd2}, and easily win the game from there, as $g(m_3,s) = w_{m_3,e_1} \cdot 0.99+w_{m_3,e_2} \cdot 0.99 = 0.99$. This shows that it is not sufficient to only evaluate the absolute best moves, but rather that one needs to evaluate all possible moves, Generally, this is infeasible, as with a reasonable information set size, regarding all moves that are legal anywhere leads to too high computational requirements. Therefore, we still restrict ourselves to a subset of the possible move space.  

Despite this fundamental problem, the idea of estimating the value of an imperfect information state by reducing it to the evaluation of perfect information moves has been successful in simulation algorithms. There, using perfect information search on sampled states has been used in games such as bridge \cite{GIB}, Skat \cite{Long2010} and Scrabble \cite{lig*Sheppard99}. In a way we follow this line of work, but instead of using sampling to approximate the expected value $g(s)$, we try to approximate it by estimating reasonable weights $w_{e,s}$ from data and to explore how they can be used for imperfect information gameplay.

\subsection{Learning Information Weightings}
\label{sec:formal_weighting}

The goal of this work is to find useful weights $w_i$ for Equation~\eqref{eq:eval} by learning a function $\F: \I \rightarrow [0,1]$ which maps each $\e_i \in \I$ to a weight $w_i = \F(e_i)$ and captures the likelihood that $\e_i$ is the \textit{true state} in the current situation. $\F$ is trained from past game data including each player's observations and the full true game state information at each move. One approach to this is to train a binary classifier, which maps the currently known set of observations $\Obs_t$ and a given position $e_{t,i}$ in the information $\I_t$ to a classification signal, which indicates the probability that this position is the correct one ($e_{t,i} = \positive_t$). We believe that such an approach has several disadvantages, due to the skewed distribution of the dataset and the high evaluation complexity, which our experiments in Section~\ref{sec:cnn} support. Instead, we tackle this problem by training a \textbf{Siamese neural network.}

\section{Siamese Networks for Information Set Weighting}
\label{sec:overview}

From each observation $\Obs_t$ we can create a total of $|\I_t| -1$ \emph{triplets} of the form $\triplet{\Obs_t}{\positive_t}{\negative_{t,i}}$, which indicate that in $\I_t$, the positive example $\positive_t$ was the true game state, while each of the negative examples $\negative_{t,i}$ was not. These triplets are used to train a Siamese neural network (shown in the middle of Figure~\ref{fig:scheme}) with the triplet loss. The goal of training for the Siamese network is to embed the inputs such that the positive example $\positive_t$ is positioned closer to the observation history $\Obs_t$ in the embedding space than the negative examples $\negative_{t,i}$.

\subsection{Siamese Neural Networks for Imperfect Information Games}
\label{sec:siamese}

Traditionally, \emph{Siamese neural networks} \cite{bromley1993signature,chicco2021siamese} are used to compare the strength of a relationship of several \textit{options} to an \textit{anchor}. One famous application is one-shot learning in image recognition \cite{koch2015siamese, schroff2015facenet}, where a network is trained to model that the \textit{positive} image $\positive$ is more similar to an \textit{anchor} image $\anchor$ than a \textit{negative} image $\negative$. 

Given triplets $\triplet{\anchor}{\positive}{\negative}$, comparisons are constructed by feeding all three items $e \in \{\anchor, \positive, \negative\}$ to a neural network $\F$, resulting in high-dimensional output embedding vectors $\F(\e)$. The parameters of $\F$ are trained such that distances between these embeddings model the strength of the relationship of the inputs, with the goal that $d_p < d_n$,  as illustrated in the centre of Figure~\ref{fig:scheme}.\footnote{For a triplet $\triplet{\anchor}{\positive}{\negative}$, given a distance metric $d$, we define $d_p = d(\F(\anchor),\F(\positive))$ and $d_n = d(\F(\anchor),\F(\negative))$, omitting the argument $a$ for brevity.} This is commonly achieved by minimising the \emph{triplet loss}. 
\begin{equation}
\label{eq:triplet}
    L_{\textrm{triplet}}(\anchor,\positive,\negative) = \max\left(d_p-d_n + m,0\right).
\end{equation}
Here, the margin $m$ controls the minimum distance difference between the positive and negative examples necessary to achieve zero loss. As shown at the top of Figure~\ref{fig:scheme}, we set $m=1$. This parameter is based on heuristic results, although we found little difference between choices. For the distance metric, we chose the Euclidean distance.

As noted above, we apply a Siamese network to model the weights of each specific board state of an information set, using the observation history as the context, with training triplets of the form $\triplet{\Obs_t}{\positive_t}{\negative_{t,i}}$.
Critically, in contrast to previously discussed tasks, the imperfect information setting usually does not possess a single \enquote{true answer}. Which board occurs in a game depends on the non-deterministic move choice of the opponent. However, our network attempts to learn from historical training data which board states are more likely, creating a probability distribution rather than deciding on a single state.

As a second key difference between the image recognition setting and our use case, all inputs to a Siamese image recognition network usually share a common vector representation. All inputs are of the same type; images. In our setting, the anchor $\Obs_t$ encodes a different type of information, the observation history, than the two boards that are compared. Therefore, we add two smaller encoding networks, shown in blue and green in Figure~\ref{fig:scheme}, which transform the board states and the observation history into matching latent encodings, which can then be used as same-type inputs for the Siamese network. More details on this are given in Section~\ref{sec:architecture}.

\begin{figure}
    \centering
    \includegraphics[width = \columnwidth]{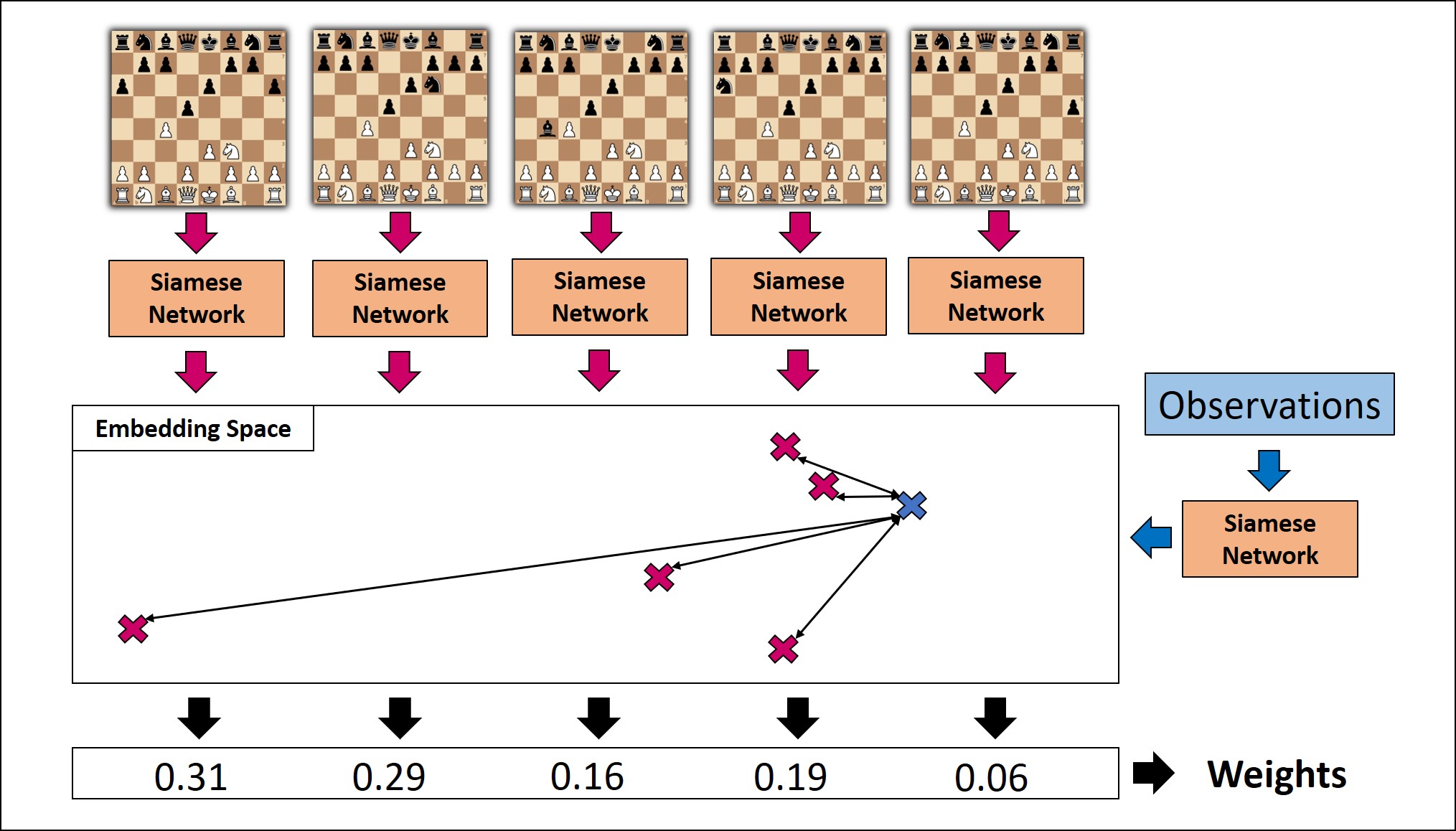}
    \caption{Process of generating a weighting over an information set. All boards and the current history of observations are fed into the Siamese Neural Network. The distances in the embedding space between the boards and the observations are computed and used to yield a weighting over the boards.}
    \label{fig:testing_scheme}
\end{figure}

The combined network produces an embedding that maps the inputs into an embedding space in which the distance between the observation and a board in the information set model the probability that the position occurs in gameplay. Thus, we can use the embedding distances as the information set weights (Figure~\ref{fig:testing_scheme}).

\subsection{Related Work on Siamese Networks in Game Playing}
\label{sec:related_work}

Siamese architectures have been used for tasks other than image recognition. However, when used for game position evaluations, they are often based on comparisons without explicitly modelling the anchor. Tesauro used such a pairwise Siamese neural network to evaluate which of two Backgammon positions should be preferred over the other \cite{tesauro1988connectionist}, and DeepChess \cite{david2016deepchess} compares chess positions with a twin Siamese neural network. Both of these applications compare fully observable board states to learn evaluations of the positions. In our work, we add a player's observation history as a context, which is required to relate the preference to a specific situation, and fundamentally changes the way the neural network is trained. Crucially, we can not compare arbitrary positions, as we can only relate states in the same information set. In addition, we do not learn evaluations, but rather their likelihood of occurrence.

One can also view each triplet as a contextual preference, denoted as $\left(\positive_t \succ \negative_{t,i} \mid \Obs_t \right)$. In this way, our approach becomes a version of \textit{contextual preference ranking (CPR)}, which uses Siamese neural networks for preference-based decision-making \cite{bertram2021predicting}. In particular, Bertram et al.\ used CPR to measure the synergy of cards in a collectable card game, where the Siamese network modelled how well a candidate card fits a set of previously chosen cards. Here, we extend this idea of relating decisions to the context in which they occurred to the case of observation histories under imperfect information.

\section{Experimental Setup}
\label{sec:experiments}

\subsection{Preparation of Gameplay Data}
\label{sec:preparation}

We obtained $582\,450$ games of \textit{RBC} as training data for our network \cite{RBC}. Each recorded game includes a turn-by-turn list of all observations received from each player's perspective, and supplementary information such as the name of the opponent. From this information, we form the anchors $\Obs_t$ shown at the bottom of Figure~\ref{fig:scheme}. As mentioned, it is possible to fully reconstruct the information set for each player and action, but for some players, mainly naive or malfunctioning ones without a reasonable sensing strategy, the information set can grow too large, so we limit them to $5\,000$ boards. In addition to the information set and the observations, we also extract the board that represents the true underlying game state. Together, the \emph{observations}, the \emph{true (positive) board}, and one \emph{wrong (negative) board} form the triplets used for training the neural network (Figure~\ref{fig:scheme}). This results in a total of 27 million samples, which are split 90/10 into training and test data to compute an out-of-sample accuracy estimate for the trained neural network. In training, we take precautions to prevent oversampling decisions with large information sets (see Section~\ref{sec:Training_process}).

\subsection{Representation of Boards and Observations}
\label{sec:sample_representation}

Each chess position is represented by a $12\times 8 \times 8$ bit tensor which encodes the occurrences of the $12$ different chess pieces. This representation omits some details, such as castling rights and turn numbers, but captures the vast majority of information.  We represent a truncated history of observations as follows: For each turn of the game, a $90 \times 8 \times 8$ bit tensor (Table~\ref{tab:observations}) encodes all information received from one player's point of view \cite{bertram2022supervised}. The majority of this encoding is taken up by specifying the last-requested move by the player \cite{silver2018general}. We truncate the observation history to the 20 most recent turns, padding the input if fewer turns were played and discarding any turns further in the past. Finally, we one-hot encode the 50 most prominent opponent's names in 50 $8 \times 8$  planes, since this information can have a large influence on the policy of an agent \cite{clark2021deep}. If the opponent's name is not in this list of players, all 50 planes are set to zero. Thus, the total observation representation is $20\times90+50 = 1850$ bit tensors of size $8\times8$.

\begin{table}\centering
\ra{1.3}
\caption{Encoding of the observations in one turn}
\label{tab:observations}
\resizebox{\columnwidth}{!}{%
\begin{tabular}{@{}ll@{}}
\toprule
\# of planes & Information represented\\
\midrule
1             & Square where the opponent captured one of our pieces \\ 
73            & last move encoded as a  $73\times 8 \times 8$ bitmap as in AlphaZero \cite{silver2018general}      \\ 
1             & Square where agent captured opponent's piece     \\ 
1             & 1 if the last move was illegal                       \\ 
6             & Position of own pieces (One plane per piece type)\\ 
1             & Last sense taken                        \\ 
6             & Result of last sense (One plane per piece type)  \\ 
1             & Color                                            \\ 
\bottomrule
\end{tabular}%
}
\end{table}

\subsection{Neural Network Architecture}
\label{sec:architecture}

\subsubsection{Encoding Networks}

To translate the two different representations of boards and observation histories into a common input to the Siamese network, we use two small convolutional neural networks that differ in input but share the same output format. In training, the boards and observations are transformed using their respective encoding network (blue and green in Figure~\ref{fig:scheme}) before reaching the common Siamese neural network (orange). They are basic convolutional neural networks with $5$ layers, $64$ filters per layer, and the ELU activation function \cite{clever2016ELU}, only differing in the shape of the input, but sharing the same inner structure and output. Each encoding network translates its input into a feature tensor with a shape of $128\times8\times8$.

\subsubsection{Siamese Network}
The Siamese neural network is a larger convolutional neural network and uses $10$ convolutional layers with $128$ filters each, also uses ELU activations, but additionally utilises skip connections. The output block of the network uses two more convolutional layers but decreases the filter size from $3\times3$ to $1\times1$ to combine the different feature planes into one final output. A single fully-connected $\tanh$ layer forms the final output of the network. We tested several architectures with more fully-connected layers, but predominantly using convolutions lead to substantially better results. The dimensionality of the resulting embedding space, i.e., the number of output neurons of the final fully-connected layer, was set as $512$, as smaller values provided insufficient ability to create the desired space.

\subsection{Training Process}
\label{sec:Training_process}

The neural network is trained on the dataset described in Section \ref{sec:sample_representation} using mini-batches of $1024$ triplets with a learning rate of $0.0001$. Larger learning rates significantly harmed the training process. For gradient updates, we use the AdamW algorithm \cite{Loshchilov2019AdamW}. The combined network, consisting of both encoding networks and the Siamese network, was trained end-to-end, driving the encoding networks towards latent representations that are most useful for the Siamese network. Pretraining the encodings networks as autoencoders did not lead to better performance.

Crucially, our training procedure differs significantly from training Siamese networks for image recognition. When training with triplets of images, it is possible to choose arbitrary combinations of images, as long as one can be considered more similar to the anchor than the other. For our task, triplets are more restricted. The preference of the positive board over all the negative boards in its information set is only valid for the specific observation in which this comparison occurred, thus vastly restricting the boards that can be related. In order to avoid oversampling from large information sets, we define one epoch of training as follows:

\begin{itemize}
    \item One \textit{epoch} consists of training with  each anchor and positive example once.
    \item For each of those pairs, one single negative board is sampled from all possible options. While it is possible to use uniform sampling, this can easily lead to generating uninformative triplets with unlikely negative boards. Instead, we choose triplets in a way that is related to semi-hard triplets \cite{schroff2015facenet}: $x$ negatives are randomly sampled from the set and their distances to the anchor are computed. For the computation of the loss and backpropagation, only the negative in the closest proximity to the anchor, i.e. the one regarded most likely, is used, thus increasing the difficulty of training. At the start of training, $x$ is set to 3 and is increased after epochs where improvements stall.
\end{itemize}

The Siamese neural network was trained with early stopping by monitoring the evaluation loss and finishing the training after 3 epochs with no improvements. In total, this process trained for 22 epochs, which took 61 hours on a single Nvidia A100 GPU. 

\section{Comparison to Conventional Neural Network}
\label{sec:cnn}

Although using pairwise comparisons to train the model is intuitive based on the raw information in the data, one can also train a classical, non-Siamese neural network similarly. We thus provide a baseline reference of performance with a classical, non-Siamese convolutional neural network. We compare both approaches, which mostly differ in the loss function used, and highlight the strength of our Siamese architecture.

\subsection{Learning RBC with a Convolutional Neural Network}

To provide a comparative baseline, we model the training of the convolutional neural network closely to the training of the Siamese network. W we use the same architecture, data, and training process as outlined in Section~\ref{sec:preparation} with the following changes:

\begin{itemize}
    \item Instead of training with triplets, we use a concatenation of the observation history $\Obs_t$ and either the true board $\positive_t$ or an incorrect board $\negative_t$, which forms a single input of shape $1862 \times 8 \times 8$.
    \item The network learns with binary classification and is trained to output $1$ if the input contains $\positive_t$ and $0$ if it contains $\negative_t$.
\end{itemize}

\subsection{Limitations}

One immediate problem with this approach is an increase in the computational complexity of weighting an entire information set. Both the Siamese and the single convolutional network evaluate the information set via a forward pass of the observation history and every game state in the state. However, the observations have to be passed through the Siamese networks only once and can then be used to compute all distances in the embedding space (Figure~\ref{fig:testing_scheme}). With the conventional architecture, observations need to be paired with every single board in the information set, unnecessarily bloating the input and causing computational overhead. 

To a lesser extent, the Siamese network architecture may also provide additional functionality than just generating weights. While we solely focus on the distances of the embeddings here, the embeddings themselves are informative and might yield more insights into the positions. Lastly, we will see in the next section that the Siamese architecture simply leads to higher accuracy than its conventional counterpart.

\section{Predictive Evaluation}
\label{sec:Evaluation}

As a first evaluation of our approach, we use the dataset from Section~\ref{sec:preparation} to test the accuracy of the network's predictions on a held-out test set. In Section~\ref{sec:playing_performance}, we add an evaluation through real game-play.

For each game position in the test set, we embed all states in the information set and the observation history with the Siamese network (see Figure \ref{fig:scheme}). In the embedding space, we rank all states in ascending order based on their distances to the observations and check the position of the true board in the ranking.

We report two metrics:
\begin{itemize}
    \item \textbf{top-k-percentage accuracy}: how often is the true board ranked in the top k-percent boards? We choose this over top-k accuracy, as the size of information sets varies greatly.
    \item \textbf{pick-distance}: the position of the true board in the ranking of all boards in the information set.
\end{itemize}

We compare our results to four baseline rankings: random ranking, ranking by the evaluations of the boards given by Stockfish, ranking by using the internal evaluation of StrangeFish2, and ranking using the probabilities of the classical CNN from Section~\ref{sec:cnn}. Using classical chess engines for ranking assumes that the opponent is more likely to play strong moves, thus leading to boards that have higher evaluations. While such evaluations cannot directly translate to \textit{RBC} evaluations, and often give vastly differing results, most current strong \textit{RBC} agents use a classical chess engine  \cite{gardner2020first,perrotta2022second}. Here, we especially regard StrangeFish2 \cite{Perrotta_StrangeFish2_2022}, the currently strongest player on the public leaderboard, which uses Stockfish with several RBC-specific adaptations. 

\begin{table}[h]\centering
\caption{Top-1 board prediction accuracy for different opponents in dataset. Opponents sorted by current elo}
\label{tab:player_accuracy}
\begin{tabularx}{\columnwidth}{@{}lrclr@{}}
\toprule
Opponent & Top-1 accuracy & \phantom{abc} & Opponent & Top-1 accuracy\\
\midrule
StrangeFish2 &\colorbox{1}{0.39} & & Stockenstein &\colorbox{2}{0.29}\\
Fianchetto &\colorbox{13}{0.56} & & Testudo &\colorbox{12}{0.38}\\
JKU-CODA &\colorbox{10}{0.41} & & genetic &\colorbox{14}{0.25}\\
Châteaux &\colorbox{16}{0.45} & & Marmot &\colorbox{11}{0.34}\\ 
Kevin &\colorbox{15}{0.54} & & Dyn. Entropy &\colorbox{6}{0.35}\\
ROOKie &\colorbox{8}{0.44} & & trout &\colorbox{7}{0.67}\\
StrangeFish &\colorbox{4}{0.46} & & attacker &\colorbox{3}{0.93} \\
Oracle &\colorbox{5}{0.50} & & random &\colorbox{2}{0.28}\\
penumbra &\colorbox{10}{0.43} & & No player &\colorbox{9}{0.39}\\
\bottomrule
\end{tabularx}
\end{table}

Table~\ref{tab:player_accuracy} shows the top-1 accuracy for frequent opponents, i.e. how often the Siamese network correctly identifies the correct board in the information set. Note that the identity of the opponents is encoded in the observation history and may therefore influence the embedding. As a result, different boards may be selected for the same information set if the opponents differ. We find that the predictability of the opponent's behaviour varies between different players. Even though \textit{random} initially might seem unpredictable, the network is able to learn that this player often makes illegal moves, thus not changing the board and allowing for some modelling. In the other extreme case, boards from games against the strongly scripted \textit{attacker} are easiest to predict, as it only follows a few fixed strategies. Altogether, the Siamese network can model all opponents to some degree, including those with the highest Elo ratings, \textit{Strangefish2}, \textit{Fianchetto}, \textit{JKU-CODA}, and \textit{Châteaux}. This is especially reassuring given that the games of these players result from dozens of different versions of different strengths that all play under the same name. Finally, we observe that in Table~\ref{tab:player_accuracy}, the left column with higher-rated opponents seems to be more consistent in predictability than the right column with lower-strength players.

\begin{figure}[t]
    \centering
    \includegraphics[width = \columnwidth]{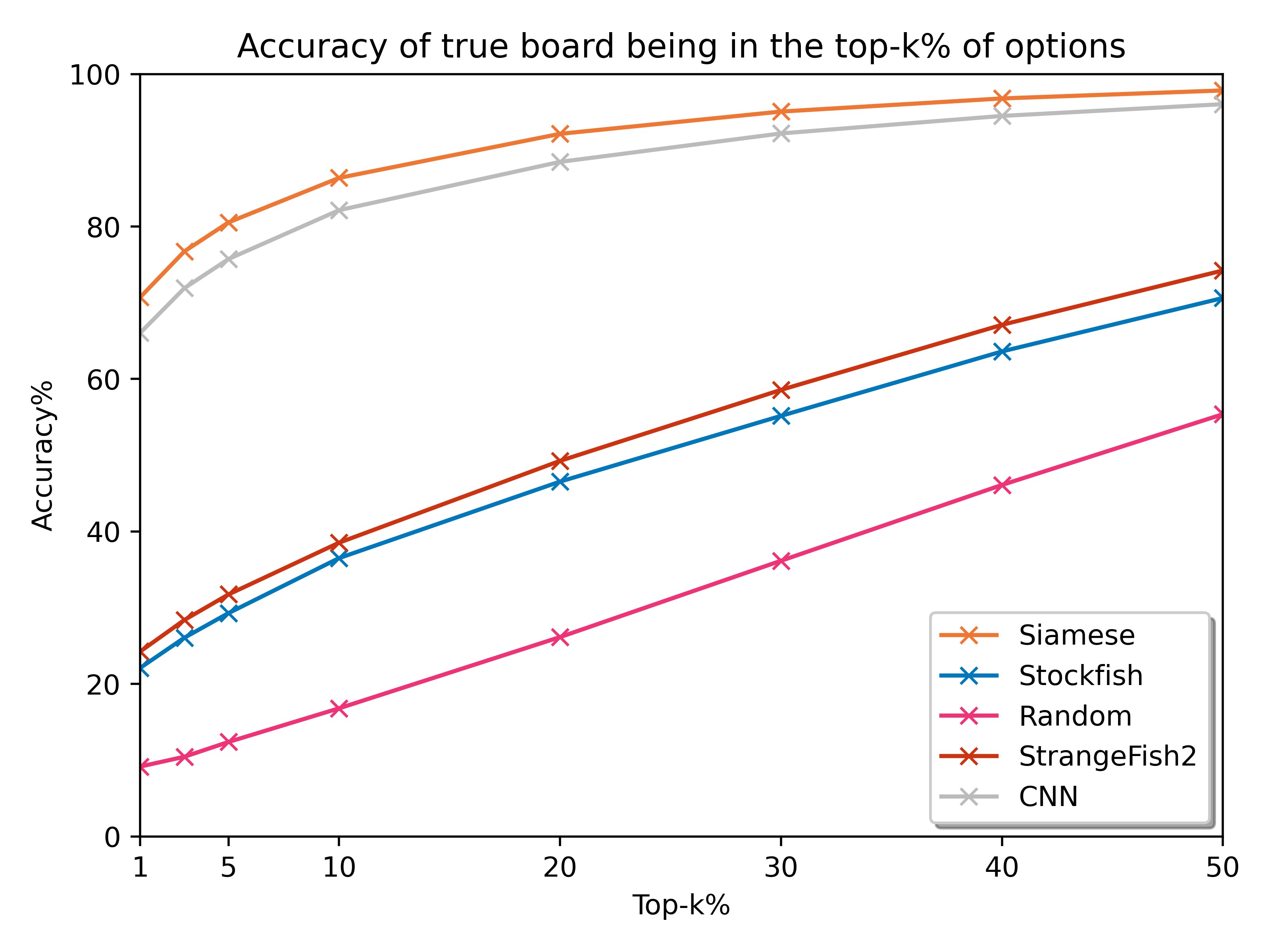}
    \caption{Comparison of top-k-percent accuracy of the Siamese network to random, Stockfish, StrangeFish2, and CNN ranking. The average size of the information set is 1100. The Siamese network vastly outperforms the three non-neural network baselines, achieving much higher accuracies than all of those metrics. The CNN performs better than the other baselines, but the Siamese network is able to create better rankings by a noticeable margin. The CNN is able to correctly identify the correct board with an accuracy of 48.82\% while the Siamese network achieves 52.91\%.}
    \label{fig:accuracies}
\end{figure}

Figure \ref{fig:accuracies} shows the top-k-percent accuracy of the five methods. Ranking boards by their Stockfish evaluation achieves higher accuracy than random ranking, and the additional adaptations of StrangeFish2 result in slight further improvements. While training a CNN to explicitly model the likelihood of a given position gives a better accuracy, the Siamese network achieves the best performance overall. In more than half of the samples (52.91\%), the true board is ranked first and in 98\% of cases, the true board is ranked in the top half of all options. 

Note that while the Siamese network is able to achieve better weightings and board predictions than the algorithm in StrangeFish2 provides, this does not necessitate that our agent plays better overall. In this experiment, we solely compare the weighting scheme itself on a held-out test-set. This does not involve gameplay and how the weighting is used in practice. We provide a gameplay-based test in Section~\ref{sec:playing_performance}.

\begin{figure}
    \centering
    \includegraphics[width = \columnwidth]{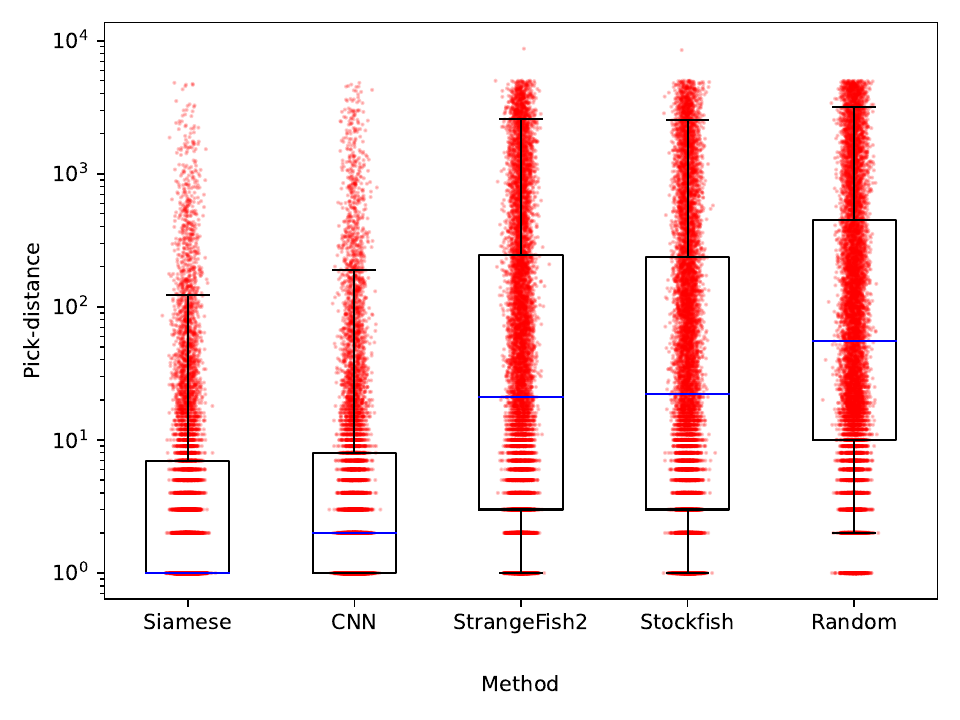}
    \caption{Comparison of pick-distance of the true board when using the Siamese network, the classical CNN, Stockfish, StrangeFish2, and random ranking. Individual samples per method have small uniform noise on the x-axis added for better visualisation. The Siamese network has some amount of outliers due to the stochastic nature of the task, but is generally able to achieve a high ranking of the correct choice and a median rank of 1.}
    \label{fig:pick_distance}
\end{figure}

Figure \ref{fig:pick_distance} provides more insights into the performance of the network by summarising the individual pick-distances for numerous samples. The Siamese ranking is clearly better at modelling player behaviour than the three non-neural network baselines, and again is more accurate than the CNN. In the majority of cases, the pick distance is below 10, and only about 5\% of samples have a distance over 100.

\section{A Siamese RBC Agent}
\label{sec:rbc_agent}

\begin{figure*}[t]
    \centering
    \includegraphics[width = \textwidth]{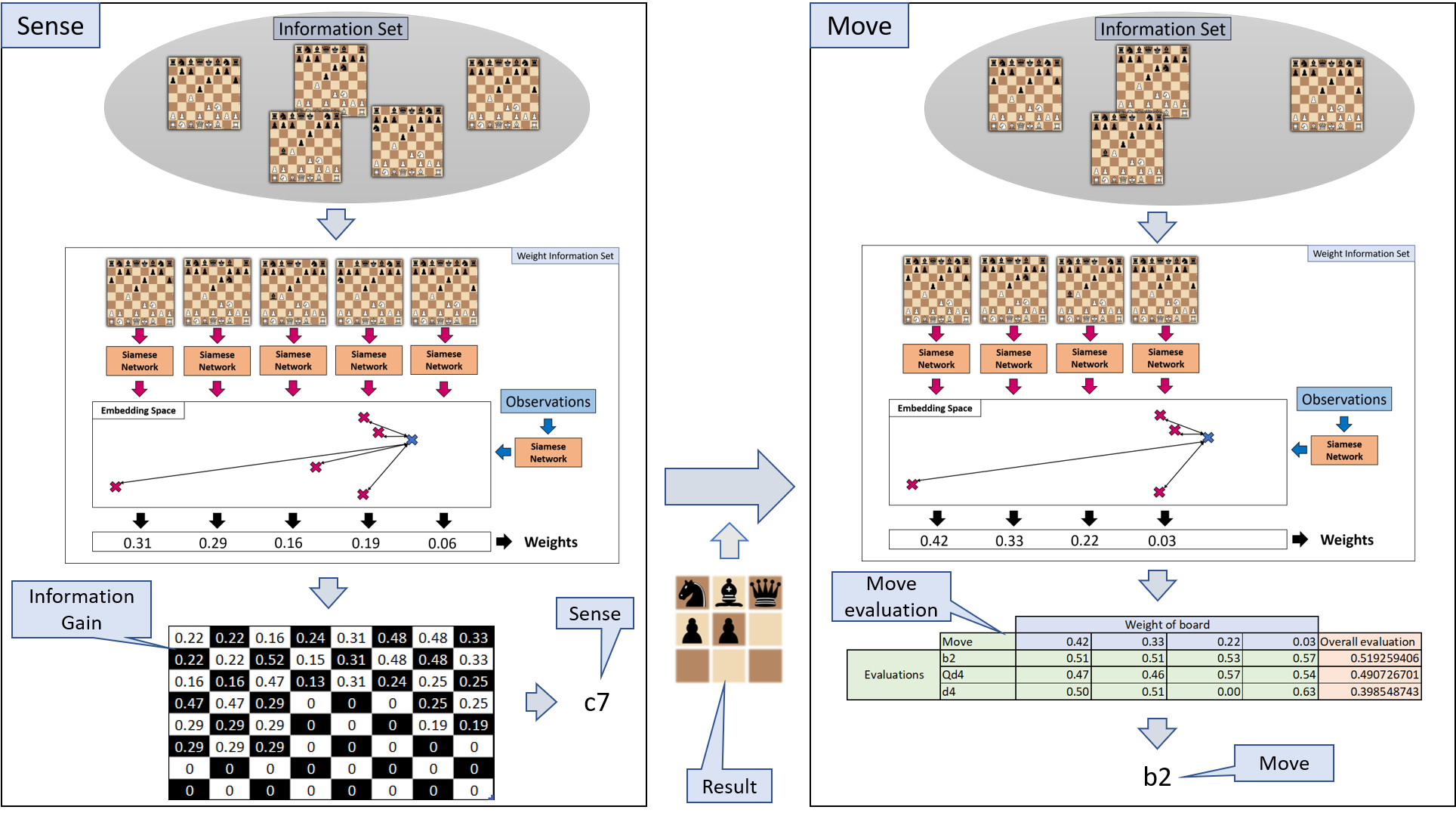}
    \caption{Overview of the \textit{RBC} agent used in the experiments. In the sensing phase, the Siamese network is used to weight the information set to maximise expected information gain. When moving, the network's weighting is utilised to identify the move with the best overall evaluation.}
    \label{fig:agent}
\end{figure*}

To see the potential gain of information set weighting, we implement and evaluate an \textit{RBC} agent that strongly utilises the Siamese neural network to aid sense and move selection. An overview of the basic concept of the player is shown in Figure~\ref{fig:agent}.

Building an \textit{RBC} agent requires three main components: (i) handling information received throughout the game, (ii) choosing a sensing action, and (iii) selecting the move to be played. Our agent is built around the idea of tracking the information set of board states, starting with the initial piece configuration and adding and removing states based on the received information. The trained Siamese network is heavily used for both sensing and moving. When sensing, the agent aims to minimise a weighted measure of the information set size, which is achieved by computing the weighted number of board conflicts per sensing square (see Section~\ref{sec:sensing}). To choose a move, the agent uses Stockfish\footnote{\url{https://stockfishchess.org/}} to evaluate the strength of a move on each board, and weights them to receive an overall evaluation of a move across the information set (see Section~\ref{sec:moving}).

\begin{algorithm}
\caption{Compute \textit{sense} evaluation of each square}
\label{alg:sense}
\KwData{boards, weights}
\KwResult{one score for each square}
$\textrm{score} \gets \textit{zeros}\,(64)$\\
$\textrm{end} \gets \min(100,\textit{len}(\textrm{boards}))$\\
\For{\emph{\textrm{square}} $\gets 0$ \KwTo $63$}{
    $\textrm{diffResults} \gets \textit{dictionary}\,()$\\
    \For{$i \gets 0$ \KwTo $\textrm{end}$}{
            $\textrm{res} \gets \textit{senseResult}\,\textrm{(boards[i],square)}$\\
        \eIf{\emph{\textrm{res}} $ \in$ \emph{\textrm{diffResults}}}{
            $\textrm{diffResults\{res\}} \gets \textrm{diffResults\{res\}} + \textrm{weights}[i]$
        }
        {
        $\textrm{diffResults\{res\}} \gets \textrm{weights}[i]$
        }
    }
    $s \gets \textit{sum}\textrm{(diffResults)}$\\
    \For{$\emph{\textrm{res}} \in  \emph{\textrm{diffResults}}$}{
    $\textrm{score[square]} \gets \textrm{score[square]} + (\textrm{res}/s)\cdot(s-\textrm{res})$\\
    }
}
\end{algorithm}

\subsection{Handling Information Sets}
\label{sec:handling_information}

The majority of strong \textit{RBC} agents are based on tracking the information set of board states $\I$ \cite{perrotta2022second}. For computing board probabilities, our agent also needs to track $\I$. After each opponent's turn, the set of possible states is replaced by all possible states that could follow each of the boards in the previous set. If the opponent captured a piece, this set decreases in size, as only a limited number of previous states allow capturing pieces, while non-capture moves greatly increase the cardinality. Whenever the agent itself senses or moves, it removes boards from the information set that are inconsistent with the new observations. Crucially, efficient sensing is important, as the information set can quickly grow unmanageably if too few states are removed.

\subsection{Choosing a Sensing Action}
\label{sec:sensing}

Most \textit{RBC} agents mainly sense to reduce the information set size \cite{perrotta2022second} and our agent follows this practice. To sense, a weight distribution over the information set is computed from the distances in the embedding space created by the Siamese network. The distances are then divided by a pre-set temperature parameter, that scales the optimism of the agent towards the weights (see Section~\ref{sec:temperature}), and fed into a Softmin to receive normalised weights. Based on this, the agent computes a score for each of the possible sensing locations.\footnote{In practice, the 28 edge squares are ignored because a superset of the information can be obtained from sensing an adjacent interior square.} The score estimates the expected number of board state eliminations based on conflicts of possible sensing results, weighted by the probability of each of the top 100 likeliest boards (see Algorithm \ref{alg:sense}). Finally, the area with the highest score is chosen.

\subsection{Choosing a Move}
\label{sec:moving}

After receiving new information from the sense, the agent appends this observation to its history, which shifts the anchor in the embedding space. Then, the agent again computes a weight distribution over the remaining boards in the information set. To decide on a move, a candidate selection has to be generated. Move candidates could simply be all possible moves, i.e. every move that is possible to play on any board, but this set quickly grows out of hand for a larger information set. Instead, we choose to generate a subset of candidates by querying Stockfish for the best move on each board in the information set. While this can lead to critical failures (see Section~\ref{sec:formal_weighting}), this served as the most principled approach in practice. To compute the best move, we individually evaluate each candidate on all states of the information set with Stockfish and compute a weighted sum of the evaluations with the Siamese network. The move with the highest overall evaluation is chosen.

\subsection{Weighting Temperature}
\label{sec:temperature}

While we can rank boards based on their distances in the embedding space, using the distances as weights first requires a transformation of them. Raw distances are arbitrary values which only possess meaning in relation to the other distances from the information set. Simply normalising them leads to a smooth weight distribution, thus  overemphasising unlikely boards. Similarly, when only using the minimum (as previously done \cite{Bertram_Weighting_Information_Sets}) or a Softmin on the unaltered distances, many reasonably likely boards are disregarded. Thus, we scale the distances with a temperature parameter $t$ before using a Softmin to receive weights. For $t = 1$, we receive the unchanged distances, which practically results in taking the minimum, while for $\lim_{t\to\infty}$ we receive a uniform weighting. We test multiple choices of this parameter in Section~\ref{sec:playing_performance}.

\section{Playing Reconnaissance Blind Chess}
\label{sec:playing_performance}

Finally, we evaluate the agent described in Section \ref{sec:rbc_agent} on the publicly available \textit{RBC} leaderboard \cite{RBC}. The \textit{RBC} leaderboard is a freely available service which provides automated testing against all other currently active agents. While active players change, a small selection of agents is always connected, whose strength varies from rather basic to state-of-the-art. This ensures that players can be assigned a representative Elo rating, albeit with some variance.

\begin{figure}
    \centering
    \includegraphics[width = \columnwidth]{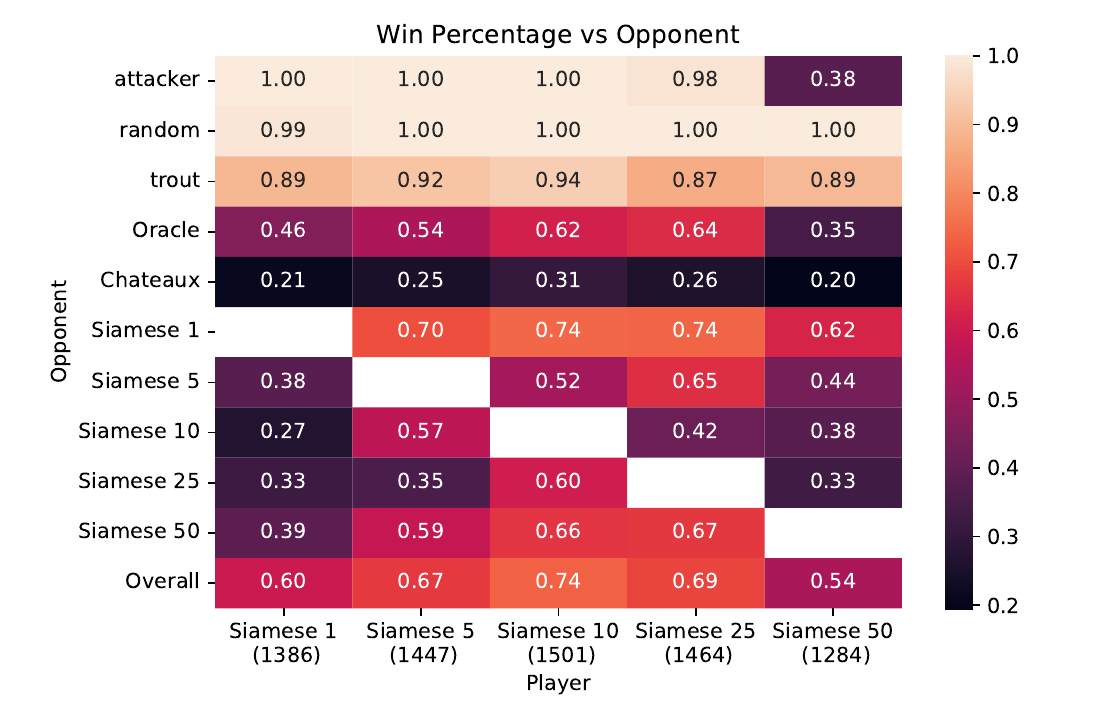}
    \caption{Win percentage of the Siamese agent with different parameter choices, where Siamese{t} stands for a denominator of t for the distances before using Softmin to compute the weights. t=1 practically corresponds to taking the minimum, while t=50 leads to an almost uniform weighting. The number in brackets shows the final ELO on the public leaderboard. We find that a moderate setting of 10 performs best, thus not only taking the most likely board into account but still strongly relying on the Siamese network.}
    \label{fig:winpercent}
\end{figure}

We previously found that using the Siamese weighting performs significantly better than the baselines \cite{Bertram_Weighting_Information_Sets}, and extend the comparison by exploring different temperature settings. The result of this test is shown in Figure~\ref{fig:winpercent}, where \textit{Siamese~\{t\}} signals that this version uses a temperature of $t$.

With Figure~\ref{fig:winpercent} we can confirm that neither a very sharp nor a very flat weight distribution leads to optimal performance. Rather, a moderate setting of $t=10$ performed best with $t=5$ and $t=25$ as the two next best settings. While sample sizes are low, the win rates against specific opponents show the influence of the temperature parameter. Against \textit{attacker}, a confident defence is crucial to success, thus the flat distribution leads to a large drop in performance. All other tested agents are able to easily exploit this opponent, but a too-high temperature hinders the predictive ability of the model. When directly comparing our agents, $t=10$ is the only one that achieves a positive win rate against the others Siamese agents, and also receives the overall best performance on the leaderboard. At the time of writing, our agent \textit{Siamese 10} is ranked \#5 on the public leaderboard.

\section{Conclusion and Future Work}

We trained a Siamese network to estimate situation-specific probabilities of states in an information set. Our experiments show that the network is able to perfectly identify the true state in many cases, even when the information set is large. In addition, we see that this allows for a perfect information approach to the imperfect information game of Reconnaissance Blind Chess. We build an agent that evaluates a set of moves on all possible boards with Stockfish and combines the resulting evaluations with the weight distribution obtained from the Siamese network. This approach prunes unlikely situations, which leads to overall better play. Thus, we reduce \textit{RBC} to evaluations of classical chess and show good performance, such that our agent is currently ranked among the strongest on the public leaderboard. 

Despite this, we consider the fact that moves are still selected using Stockfish to be the main weakness of our current approach. In particular, RBC-specific policies, such as strategies using speculative attacks on the opponent's king, can not be generated in a setting where moves are obtained from a classical chess engine. Changing the final move selection, for example, to an RBC-specific neural network, or conducting a search with our learned probability distributions, may further improve our agent. 

Although only tested in one domain so far, our underlying method is general and can be used in other applications for weighting information sets. We intend to further explore future use in other imperfect information decision-making tasks.
\newpage


%


\section*{Acknowledgments}

M\"uller acknowledges financial support from NSERC, the Natural Sciences and Engineering Research Council of Canada, DeepMind, and the Canada CIFAR AI Chair program.
We are grateful to Ryan W. Gardner and the John Hopkins University for organizing the RBC competitions and maintaining the ladder, as well as to the authors of open source programs such as Strangefish and Stockfish that we could use in our work.

\ifCLASSOPTIONcaptionsoff
  \newpage
\fi



%

\bibliographystyle{IEEEtranS}
\bibliography{bibtex/bib/IEEEabrv, bibtex/bib/bib}

%

\begin{IEEEbiography}[{\includegraphics[width=1in,height=1.25in,clip,keepaspectratio]{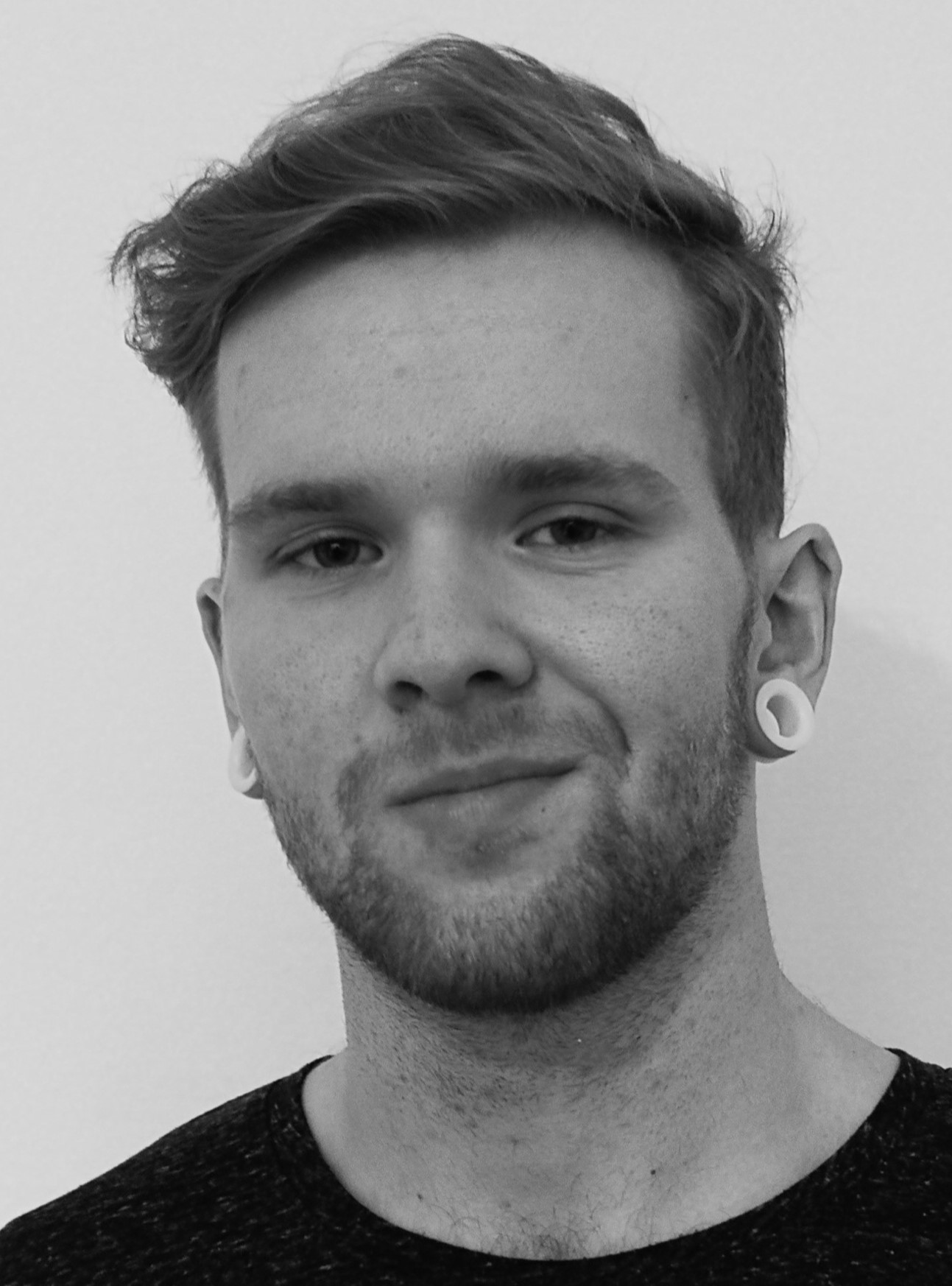}}]{Timo Bertram} is currently working towards the Ph.D. degree in Artificial Intelligence at the Johannes Kepler University in Linz, Austria. He obtained an M.Sc. in computer science from the University of Birmingham, UK, and a B.Sc. in informatics from the University of Bonn, Germany. His research focuses on artificial intelligence for games, mainly using contrastive learning and preferences to train agents and model player behaviour.
\end{IEEEbiography}

\begin{IEEEbiography}[{\includegraphics[width=1in,height=1.25in,clip,keepaspectratio]{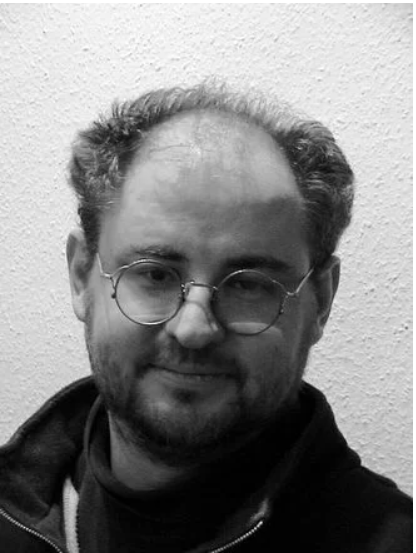}}]{Johannes F\"urnkranz}
is Professor for Computational Data Analytics at the Johannes-Kepler University in Linz, Austria. From 2004--2019, he was Professor for Knowledge Engineering at TU Darmstadt, Germany. 
He obtained Master Degrees from the
Technical University of Vienna and the University of Chicago, and a
Ph.D. from the Technical University of Vienna.
His main research interests are machine learning and data mining, in
particular inductive rule learning and interpretable models, as well as
their applications in game playing. 
He served as the editor-in-chief of the journal \emph{Data Mining and Knowledge Discovery} (2015--2023), and as the PC chair of several reputable scientific conferences, including ICML (2010) and ECML/PKDD (2006).
\end{IEEEbiography}


\begin{IEEEbiography}[{\includegraphics[width=1in,height=1.25in,clip,keepaspectratio]{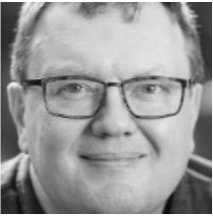}}]{Martin M\"uller}
is a professor in the Department of Computing Science, an Amii Fellow, and the DeepMind Chair in Artificial Intelligence, all at University of Alberta, and a Canada CIFAR AI Chair. He obtained a Dipl.Ing. degree from TU Graz and a Ph.D. from ETH Zurich. His main area of research is modern heuristic search, with its complex interactions between search, knowledge, simulations, and machine learning. Application areas include game tree search, domain-independent planning, combinatorial games, and boolean satisfiability (SAT) solving. M\"uller has worked on computer Go for thirty years. His group developed the open source program Fuego, the first to win a 9×9 Go game on even terms against a top-ranked professional human player in 2009. With his students and colleagues, M\"uller has developed a series of successful game-playing programs, planning systems, and SAT solvers.
\end{IEEEbiography}




\end{document}